\documentclass{article}
\usepackage[table]{xcolor}

\usepackage[preprint]{corl_2026} 

\usepackage[T1]{fontenc}
\usepackage{algorithm}
\usepackage{algpseudocode} 
\usepackage{upgreek}
\usepackage{multicol}
\usepackage{dsfont,setspace}
\usepackage{amsmath,graphicx}
\usepackage{float}
\usepackage{amssymb}
\usepackage{booktabs}
\usepackage[normalem]{ulem}
\usepackage[thinc]{esdiff}
\DeclareMathOperator*{\argmax}{arg\,max}
\DeclareMathOperator*{\argmin}{arg\,min}
\usepackage{lineno}
\usepackage{epstopdf}
\usepackage{dblfloatfix}
\usepackage{tabularx}
\usepackage{soul}
\usepackage{wrapfig}
\usepackage{tikz}
\usetikzlibrary{positioning,fit,backgrounds}

\usepackage[
  separate-uncertainty = true,
  multi-part-units = repeat
]{siunitx}

\usepackage{subcaption}
\usepackage{cleveref}
\usepackage{textcomp}
\usepackage{gensymb}
\usepackage{pifont}
\usepackage{makecell}
\usepackage{svg}
\usepackage{multirow}
\usepackage{enumitem}

\title{Shield-Loco: Shielding Locomotion Policies with Predictive Safety Filtering}

\def\AND{%
  \end{tabular}\hfil\linebreak[4]\hfil%
  \begin{tabular}[t]{c}\bf\rule{\z@}{3pt}\ignorespaces%
}
\def\And{%
  \end{tabular}\hfil\linebreak[4]\hfil%
  \begin{tabular}[t]{c}\bf\rule{\z@}{3pt}\ignorespaces%
}

\author{
  Aditya Shirwatkar$^{1,2}$, Sebastian Sanokowski$^{2}$, Shishir Kolathaya$^{1,3}$,\\
  \textbf{Aaron Johnson$^{4,5}$, Majid Khadiv$^{2}$}
  \AND
  \small \textnormal{$^{1}$Robert Bosch Center for}\\
  \small \textnormal{Cyber Physical Systems,}\\
  \small Indian Institute of Science,\\ 
  \small Bangalore, India\\
  \And
  \small \textnormal{$^{2}$Munich Institute of Robotics and}\\ 
  \small \textnormal{Machine Intelligence (MIRMI),}\\
  \small Technical University of Munich,\\
  \small Munich, Germany\\
  \And
  \small \textnormal{$^{3}$Department of Computer}\\ 
  \small \textnormal{Science \& Automation,}\\
  \small Indian Institute of Science,\\
  \small Bangalore, India\\
  \And
  \small \textnormal{$^{4}$Department of}\\ 
  \small \textnormal{Mechanical Engineering,}\\ 
  \small Carnegie Mellon University,\\
  \small Pittsburgh, PA, USA\\
  \And
  \small \textnormal{$^{5}$Institute for Advanced Study,}\\ 
  \small Technical University of Munich,\\
  \small Garching, Germany\\
  \And
  \small \texttt{\{adityasr,shishirk\}@iisc.ac.in, amj1@andrew.cmu.edu,}\\
  \small \texttt{\{sebastian.sanokowski,majid.khadiv\}@tum.de}
}

\begin{document}
\maketitle

\vspace{-15pt}
\begin{abstract}
Reinforcement learning (RL) policies enable dynamic legged locomotion but lack mechanisms to avoid violations of safety constraints that are absent during training. Large-scale offline safe learning is impractical for covering all edge cases. Existing safety frameworks either rely on reduced-order models that cannot reason about whole-body behaviors or require conservative recovery controllers that degrade task performance. We propose a predictive safety filter that post-hoc filters the nominal contact locations fed to the RL policy. When a collision is predicted, a sampling-based optimizer asynchronously searches for safer contact sequences using a full-physics model, while a learned value function bootstraps long-horizon returns. Our three algorithmic components (geometric projection of sampled contacts, momentum-augmented updates, and replica-exchange) make the optimization tractable in a discontinuous contact landscape. We validate the filter on a quadruped robot in dense, cluttered environments, both in simulation and in the real world, showing substantial reductions in safety violations with minimal deviation from the nominal input. A summary of the results can be found \href{https://www.youtube.com/watch?v=2gBCMw4LD3c}{here}.
\end{abstract}

\keywords{Legged Robots, Safety Filters, Sampling Based Optimization}


\section{Introduction}
\label{sec:intro}

Contact-conditioned reinforcement learning (RL) policies have shown dynamic locomotion across diverse morphologies \cite{omar2025learningactcontactunified, lin2025simtorealreinforcementlearningvisionbased, ciebielski2025contactconditionedlearningmultigaitlocomotion, zhang2024wococolearningwholebodyhumanoid}.
By accepting desired contact locations as input and mapping them to low-level joint commands, these policies decouple high-level contact planning from PD target or torque generation.
These policies demonstrate strong task performance, but lack an active mechanism to prevent safety violations in the environment.
This is especially problematic when deployment introduces obstacles or spatial constraints that were not present during training.

Large-scale training or finetuning of the RL policy with safety objectives \cite{yang2022safereinforcementlearninglegged} is impractical for all possible cases. 
An alternative is a \emph{safety filter} \cite{Wabersich2023DataDrivenSF}: a post-hoc layer that intervenes when a violation is imminent.
For example, Hamilton-Jacobi (HJ) reachability \cite{bansal2017hamilton, mitchell2005ATO}, Control Barrier Functions (CBF) \cite{ames2017controlbarrierfunction, ames2019controlbarrierfunction}, and Model Predictive Shielding (MPS) \cite{mohammedundefined, bastani2021, zhang2019} provide rigorous guarantees. 
However, they scale poorly to whole-body models, require hand-crafted certificates that fail under unmodeled contact, or resort to conservative recovery controllers that sacrifice performance (see Sec. \ref{sec:related}).

Hence, we propose a shielding method that addresses MPS conservatism while maintaining task objectives by operating on \emph{contact locations}.
Our key insight is that these serve as a useful kinematic abstraction for enforcing contact-related safety constraints, such as whole-body collision avoidance, in legged robots.
For a quadruped, this means optimizing over desired foot placements, and the same principle applies to humanoids or manipulators (which we leave for future work). 
However, most existing safety filters that exploit this contact structure use reduced-order models, whereas we employ a full-physics simulator to enable whole-body reasoning without sacrificing model fidelity.

Our choice for this is motivated by three factors. 
First, whole-body collisions during locomotion are primarily determined by where the feet land, so redirecting footsteps indirectly steers the whole body around or over obstacles.
Second, contact locations constitute a plan that the policy smoothly tracks over time. 
This allows the safety optimizer to run asynchronously at a lower frequency. 
This decoupling is essential for deploying very long and expensive, full-physics rollouts online, which is not possible with direct torque control.
Third, although the dimensionality equals that of joint torques in a 12DOF quadruped, the effective search space is far smaller; contact locations are physically interpretable, and the policy provides a smooth learned mapping from contacts to joint targets.
\looseness-10

At each control step, our filter rolls out the nominal contact plan from a planner or command input.
If a safety violation is predicted within a receding horizon, a warm-started sampling-based optimizer searches for safer alternative contact sequences that maximize a finite-horizon return bootstrapped by a learned value function and remain close to the nominal plan.
Further, we make the optimization tractable through three components: (i) a geometric projection of sampled contact locations onto the feasible set, (ii) a momentum-augmented update \cite{khan2023bayesian} that accelerates convergence, and (iii) a Replica Exchange \cite{dong2021replicaexchangenonconvexoptimization} strategy that maintains exploration and escapes shallow local minima.

To summarize, our contributions are:
\begin{itemize}[noitemsep, topsep=-\parskip, partopsep=0pt]
    \item A predictive safety filter operating on contact locations fed to RL policies, enabling whole-body collision avoidance without policy modification.
    \item A sampling-based optimization scheme with three tractability-enabling components, namely geometric projection, momentum-augmented updates, and replica exchange that improve the performance of the sampling-based predictive controller.
    \item Simulation and real-world validation on a quadruped, demonstrating substantial reductions in safety violations over long horizons in nonsmooth contact landscapes.
\end{itemize}



\section{Related Works}
\label{sec:related}


\textbf{Sampling‑Based Zero‑Order Optimization:}
Zero‑order optimizers such as Model Predictive Path Integral (MPPI) \cite{williams2015modelpredictivepathintegral} and the Cross‑Entropy Method (CEM) \cite{de2005tutorial} have been applied to legged locomotion for full‑order torque‑level control \cite{xue2024fullordersamplingbasedmpctorquelevel}, combined with learned value functions for planning \cite{Crestaz2026TDCDMPPITC}, and learned dynamics \cite{shirwatkar2024piplocoproprioceptiveinfinitehorizon}.
These methods handle discontinuous cost landscapes and contact dynamics well, but they require extensive sampling and iterations in joint space for real-time long-horizon safety filtering.

\textbf{Hamilton-Jacobi Reachability:}
Hamilton-Jacobi (HJ) reachability provides rigorous guarantees by computing an offline value function whose sub-zero level set defines unsafe states \cite{bansal2017hamilton}.
While effective for low-dimensional systems, it scales exponentially with state dimension, making whole-body legged models intractable \cite{yang2021reachability, borquez2023hamilton, xia2024gait}.
Moreover, the value function must be recomputed for new scenario changes, precluding online adaptation.

\textbf{Control Barrier Functions:}
CBFs enforce forward invariance via quadratic programming \cite{ames2017controlbarrierfunction, ames2019controlbarrierfunction}, and have been applied to legged robots using reduced-order models \cite{grandia2021multilayeredsafetyleggedrobots, cohen2024safety, Bena2025GeometryAwarePS, peng2024realtimesafebipedalrobot}.
However, constructing valid CBFs for intermittent multi-contact dynamics remains exceptionally difficult, and the guarantee relies on accurate model knowledge that unmodeled contacts in reduced-order models can easily violate, rendering the formal guarantees infeasible in practice.

\textbf{Predictive Safety Filters / Model Predictive Shielding:}
MPS forward‑simulates candidate actions and overrides the nominal policy with a recovery controller when a safety violation is predicted \cite{mohammedundefined,bastani2021,zhang2019}.
MPS avoids the curse of dimensionality and does not require an analytical barrier function, but its main limitation is the conservatism of standard recovery policies, which are safe but task‑agnostic.
Dynamic MPS \cite{banerjee2024} mitigates this by planning over a receding horizon with a learned value function that captures long‑term returns \cite{lowrey2019planonlinelearnoffline,sikchi2021learningoffpolicyonlineplanning}.
For legged systems, prior work has exploited contact-location abstraction within reduced-order models \cite{gu2024robustlocomotionbylogicperturbationresilientbipedallocomotion}, or learned game-theoretic fallbacks that switch policies online \cite{nguyen2025gameplayfiltersrobustzeroshot}; however, these either lack whole-body reasoning or perform only binary switching without online optimization.
\cite{pua2024safe} casts safety filtering in a nonlinear model predictive control (MPC) but requires a recoverable (viable) set \cite{wieber2008viability,yeganegi2021robust}, which is exceedingly hard to characterize for general locomotion problems.
\begin{wrapfigure}{R}{0.45\columnwidth}  
    \centering
    \includegraphics[width=\linewidth]{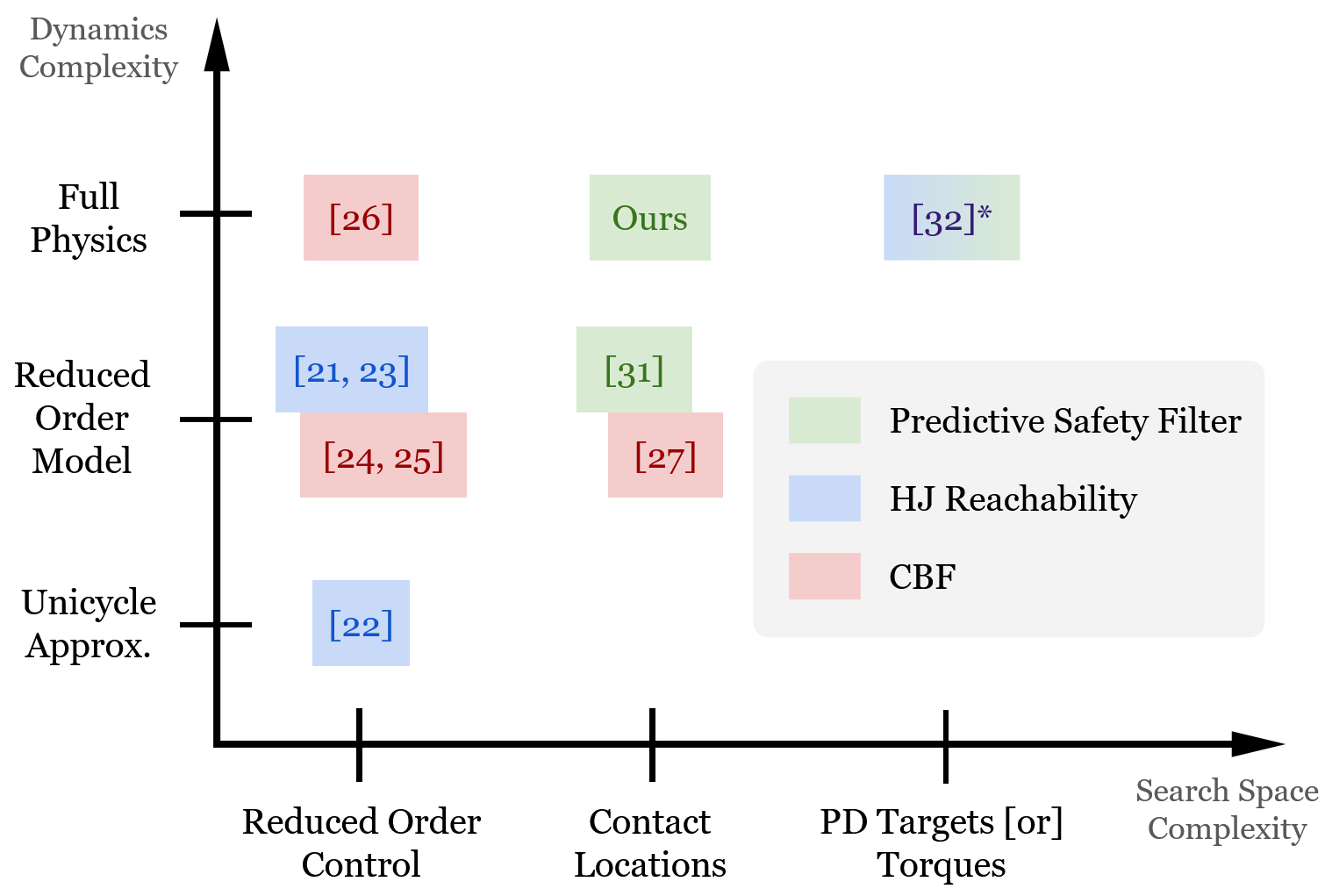}
    \caption{Landscape of safety filters for legged locomotion across search-space complexity and model fidelity. *\cite{nguyen2025gameplayfiltersrobustzeroshot} performs runtime policy switching only, without online optimization.}
    \label{fig:positioning}
\end{wrapfigure}

\textbf{Safe Reinforcement Learning with Formal Guarantees:}
Several works integrate safety into RL training while providing formal guarantees. \cite{Hsu_Ren_Nguyen_Majumdar_Fisac_2024} combines a task policy with a backup safety policy trained via the safety Bellman equation (derived from HJ reachability) and uses a shielding mechanism.
\cite{he2024agilesafelearningcollisionfree} learns an agile policy and a recovery policy whose switching is governed by a reach‑avoid value network rooted in HJ theory.
\cite{lin2024filterdeployallrobust} predicts a control‑theoretic safety value function online from LiDAR observations to construct an adaptive safety filter for quadrupedal navigation.
While such methods achieve strong safety performance within their training distributions, they remain vulnerable to out‑of‑distribution scenarios.


\textbf{Positioning:}
Figure~\ref{fig:positioning} contextualizes our approach within the landscape of safety filters for legged robots.
Rather than relying on a task‑agnostic recovery policy as in standard MPS, we re‑cast the safety intervention to be minimally invasive, i.e., deviate minimally from the nominal/desired behavior.
Furthermore, as motivated in Sec. \ref{sec:intro}, contact locations offer a principled middle ground to intervene on: expressive enough for whole-body reasoning, yet tractable enough for online sampling-based optimization.
We now detail our safety filter framework that instantiates this design choice in the next section.


\section{Methodology}
\label{sec:method}

\subsection{Preliminaries}
\label{sec:prelim}
We consider a quadruped robot controlled by a contact‑conditioned locomotion policy $\pi$ \cite{omar2025learningactcontactunified} that maps the current state and desired foot placements to low‑level joint commands. 
Let $s_t \in \mathbb{R}^{n_s}$ denote the robot's proprioceptive observation at discrete time $t$; it includes joint angles, joint velocities, the body's linear and angular velocity (in the base frame), and the 3D Cartesian positions of the four feet relative to the base. 
The policy outputs an action $a_t = \pi(s_t, p_t)$, where $p_t \in \mathbb{R}^{12}$ is the vector of desired foot contact locations (one 3D point per foot). 
In our implementation, $a_t$ is a vector of target joint positions, which are tracked by a PD controller producing joint torques. 
The closed‑loop transition dynamics are given by $s_{t+1} = f(s_t, a_t)$, where $f$ is the full physics simulation provided by the MuJoCo MJX solver \cite{todorov2012mujoco}. 
This high‑fidelity model eliminates the approximation errors typical of simplified reduced‑order models.

The locomotion reward $r(s_t, a_t)$ encodes the high‑level task (e.g., tracking a desired contact location while penalizing energy consumption) and is identical to the reward used to train $\pi$ \cite{omar2025learningactcontactunified}.
A high‑level planner produces a sequence of nominal contact locations $\{\bar{p}_{t+\tau}\}_{\tau=0}^{H-1}$ for a horizon $H$. 
However, neither $\pi$ nor the nominal contact plan may have a mechanism to actively enforce safety constraints. Hence, we introduce our general safety framework in the next section.

\subsection{General Framework}
\label{sec:framework}

\begin{figure}[t]
    \centering
    \includegraphics[width=0.9\linewidth]{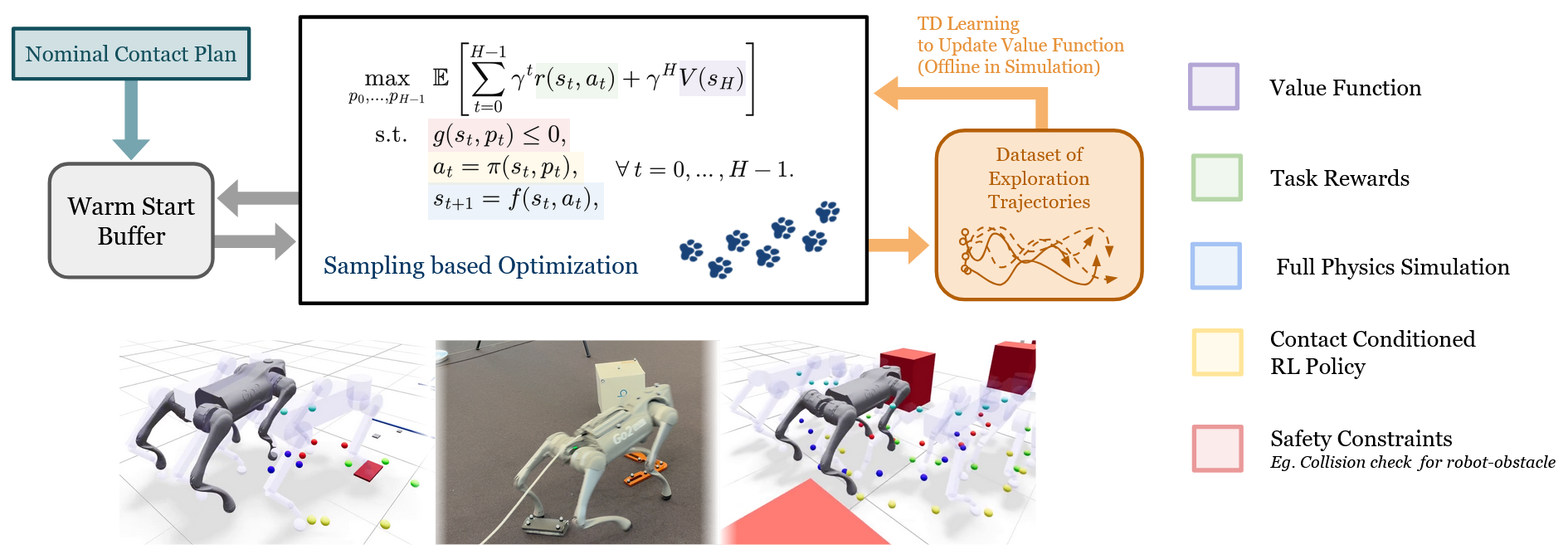}
    \vspace{-.5em}
    \caption{Predictive safety filtering pipeline. Unsafe nominal contact plans are refined through sampling-based optimization with parallel full-physics rollouts.}
    \label{fig:framework}
\end{figure}

Our safety filter is agnostic to the internals of $\pi$ and requires only that the policy accepts contact locations as input. 
At each control step, given the current state $s_t$ and a sequence of nominal contact locations $\{\bar{p}_{t+\tau}\}_{\tau=0}^{H-1}$ from the planner, the filter performs the following steps:
\begin{enumerate}
    \item \textbf{Predict:} Roll out the nominal sequence using the full dynamics $f$ and the policy $\pi$ over the horizon $H$. Check whether any safety constraint (Sec.~\ref{sec:safety}) would be violated.
    \item \textbf{Intervene only if needed:} If no violation is detected, the nominal contact $\bar{p}_t$ is sent directly to $\pi$. If a violation is predicted, the filter activates a sampling‑based optimizer that searches for an alternative contact sequence $\{p_t^*,\dots,p_{t+H-1}^*\}$.
    \item \textbf{Optimize:} The optimizer maximizes a finite-horizon objective that balances task reward and safety, using the nominal sequence as a warm start \eqref{eq:opt_obj}, Fig.~\ref{fig:framework}.
    \item \textbf{Execute:} The first optimized contact location ($p_t^*$) is fed to $\pi$, which produces joint-level actions $a_t$. The process repeats at the next time step in a receding horizon fashion.
\end{enumerate}

Formally, the optimization problem solved when the intervention occurs is:
\begin{equation}
\label{eq:opt_obj}
\begin{aligned}
\max_{p_0, \dots, p_{H-1}} \; & \mathbb{E} \left[ \sum_{t=0}^{H-1} \gamma^t r(s_t, a_t) + \gamma^H V(s_H) \right] \\
\text{s.t.} \quad & g(s_t, p_t) \leq 0, 
\quad
a_t = \pi(s_t, p_t),
\quad
s_{t+1} = f(s_t, a_t), \quad \forall\, t = 0, \dots, H-1.
\end{aligned}    
\end{equation}
where $g$ encodes safety constraints.
$V$ is the value function that bootstraps the finite-horizon return. 
We employ a scheme similar to \cite{Crestaz2026TDCDMPPITC}, where $V$ is trained offline via temporal‑difference (TD) learning
on trajectories collected by the sampling-based optimizer in simulation (more details are provided in the Appendix \ref{app:value_training}).

\subsection{Safety Constraints, \texorpdfstring{$g$}{g}}
\label{sec:safety}
We evaluate two complementary safety constraints at each forward rollout and integrate them as soft penalties. 
Crucially, the penalty depends on the actual foot positions observed in the simulated state (denoted $h_t$), not on the commanded $p_t$, thereby capturing the true closed‑loop
outcome. 
The filter assumes that the geometry and poses of all obstacles are known a priori (e.g., from a map), and in our simulation experiments, they are perfectly known. 
The extension to the online perception with uncertainty is left for future work.

\textbf{{Foot Contact Penalty:}}
All obstacles are modeled as geometric primitives (spheres, cylinders, boxes, or capsules). 
For each foot position $h_t$ (a 3D point per foot extracted from $s_t$) and obstacle $j$, we compute the signed distance function (SDF) $\mathcal{D}_j$, defined in closed form for each geometric primitive (see Appendix \ref{app:projection}). 
Then the soft penalty for that foot is given by $d_t = \sum_j \max(m - \mathcal{D}_j(h_t),\,0)^{\kappa}$, where $m$ is a safety margin and $\kappa$ controls the steepness.

\textbf{{Whole-Body Collision:}}
We query MuJoCo's contact data for every simulation step and extract a binary indicator $\mathbf{1}_{\mathrm{col}}(s_t) \in \{0,1\}$ that equals $1$ if any robot link contacts an obstacle. 
Trajectories that trigger this indicator receive a large penalty, ensuring whole-body collisions are likely to be rejected. 
Thus, the objective function from \eqref{eq:opt_obj} with relaxed safety constraints becomes:
\begin{equation}
\label{eq:total_obj}
\mathcal{S} := 
\sum_{t=0}^{H-1} \gamma^t \Big( r(s_t, a_t) - \big(
    d_t
    + M \, \mathbf{1}_{\mathrm{col}}(s_t)
\big) \Big) + \gamma^H V(s_H)
\end{equation}
where $M \ggg 0$ is chosen large enough to strongly reject colliding trajectories. 

\subsection{Sampling-Based Optimization}
\label{sec:mppi}
We solve \eqref{eq:total_obj} with MPPI, which is a derivative‑free, sampling‑based optimizer. 
These are well‑suited to the non‑smooth, discontinuous cost landscape induced by contact dynamics.
In our approach, we approximate the optimal solution by iteratively updating the contact location sequence $\mu_t \triangleq (p_t, \ldots, p_{t+H-1})$.
At each iteration, $K$ perturbations $\{\xi_t^{(k)}\}_{k=1}^{K}$ are drawn from $\mathcal{N}(0, \Sigma_t)$ to form candidates $\mu_t^{(k)} = \mu_t + \xi_t^{(k)}$.

\textbf{Projection:}
Prior to rollout evaluation, each candidate contact location is projected onto the collision-free set $\mathcal{F} \triangleq \bigl\{\mathbf{p} \mid \mathcal{D}_j(\mathbf{p}) \geq \varepsilon_{\text{safe}}, \ \forall\, j \bigr\}$ via the projection operator $\tilde{p}_t^{(k)} = \Pi_{\mathcal{F}}\big(p_t^{(k)}\big)$, yielding the candidate sequence $\tilde{\mu}_t^{(k)} \triangleq (\tilde{p}_t^{(k)}, \ldots, \tilde{p}_{t+H-1}^{(k)})$.
We solve this projection as a Quadratic Program (QP) by linearizing each SDF constraint about the current candidate, as the safety penalty in \eqref{eq:total_obj} alone does not guarantee constraint satisfaction.
Further details and an analysis of the projection's effect on solution quality are provided in the Appendix \ref{app:ablations}.

Then, for each projected candidate $\tilde{\mu}_t^{(k)}$, we roll out the dynamics to compute the weights and update the sequence with an exponentially weighted average:
\begin{equation}
    \quad w_k = \exp\!\left(\lambda \Big( \mathcal{S}^{(k)}  - \max_i \mathcal{S}^{(i)}\Big)\right), \quad \mu_{t}^\prime = \frac{\sum_{k=1}^{K} w_k \, \tilde{\mu}^{(k)}_t}{\sum_{k=1}^{K} w_k},
\end{equation}
where $\lambda$ is the inverse temperature. 
We stabilize the weights $w_k$ by subtracting the batch maximum.
It is important to note that, since the projection is only on the sampled contact targets, the actual foot locations in the rollout may violate safety constraints due to poor tracking by the RL policy.

\textbf{{Momentum-based Updates:}}
To accelerate convergence, we propose using momentum-augmented updates derived from an Accelerated Proximal Natural Gradient Descent \cite{khan2023bayesian} formulation. 
Further, these updates can be easily combined with CEM- and MPPI-like weighting $w_k$ 
Thus, using this formulation, the update for the contact sequence becomes:
\begin{equation}
\mu_{t+1} = \mu_t^\prime + \frac{\beta}{1-\beta} \Big(\mu_t^\prime - \mu_t \Big),
\label{eq:mu_update}
\end{equation}
and the updates for the covariance are computed with:
\begin{equation}
\begin{aligned}
S_t &= \sum_{k=1}^K w_k (\tilde{\mu}^{(k)}_t - \mu_{t+1})(\tilde{\mu}^{(k)}_t - \mu_{t+1})^T,\quad
\Sigma_{t+1} = S_t + \frac{\beta  (\Sigma_t - S_t) + 2 \alpha S_t}{\beta + 1 - 2\alpha} .
\label{eq:sigma_update}
\end{aligned}
\end{equation}
where $\beta \in [0, 1)$ is the momentum factor and $0\leq \alpha < \frac{\beta+1}{2}$ is an entropy coefficient.
Importantly, $\beta \in (0,1)$ yields accelerated updates that extrapolate in the direction of improvement. 
We provide a detailed derivation for Eqs.~\eqref{eq:mu_update} and~\eqref{eq:sigma_update} in the Appendix \ref{app:derivation}.

\textbf{{Replica Exchange for Exploration:}}
To avoid premature convergence to poor local optima, we employ a replica‑exchange (parallel tempering) strategy \cite{dong2021replicaexchangenonconvexoptimization}. 
We maintain $L$ MPPI replicas with inverse temperatures $\lambda_1 < \lambda_2 < \dots < \lambda_L$.
After each optimizer iteration, adjacent replica pairs $(i, \, i+1)$ swap their contact-sequences according to the Metropolis–Hastings criterion:
\begin{equation}
\label{eq:replica_swap}
\Pr(i \leftrightarrow j) = \min\!\left(1,\; \exp\!\Big[ (\bar{\mathcal{S}}_i - \bar{\mathcal{S}}_j)\big(\lambda_j - \lambda_i\big) \Big] \right),
\end{equation}
where $\bar{\mathcal{S}}_i$ is the average total return from \eqref{eq:total_obj} of replica $i$'s sample set. 
Samples themselves are discarded after each swap, as they are temperature-specific. 
This allows high-temperature replicas to explore broadly while low-temperature replicas exploit promising solutions, enabling escape from shallow local minima.

Additionally, at each new planning step, we shift the previously optimized contact sequence forward by one time step (dropping the first element) and append a nominal contact at the end. 
This shifted sequence serves as a warm start for the optimizer when intervention is required, thereby promoting temporal consistency. 
The complete safety filter is summarized as Algorithm \ref{alg:filter} in the Appendix.



\section{Experiments}
We evaluate our predictive safety filter on whole-body collision-avoidance tasks for a Unitree Go2 quadruped in the MuJoCo physics simulator and the real world. 
The filter is deployed at a planning frequency of $3\,\mathrm{Hz}$, with $H=5$ footsteps (effectively $\sim$150 full physics update steps into the future), $K=512$, and $N=3$ iterations per planning cycle. 
All experiments are run on a single NVIDIA RTX 3090, with rollouts accounting for the bulk of the computation.
The locomotion policy is trained offline using the procedure described in \cite{omar2025learningactcontactunified} and remains frozen throughout while running at $50\, \mathrm{Hz}$. 
In all experiments, the nominal planner produces footstep targets using a trotting-gait heuristic driven by a user input. 
Additional implementation details are in the Appendix \ref{app:hyperparams}.
\looseness-15

\textbf{{Scenarios:}} 
Our tests consider a dense, cluttered environment consisting of obstacles of different categories and dimensions that mimic real-world objects such as phones, packages, and cables, on which the robot must not step.
We also qualitatively test our proposed safety filter for navigating large obstacles to reach the goal and for simple dynamic collision avoidance (see Appendix \ref{app:qualitative}).

\textbf{{Baselines:}}
We compare to a \emph{CBF} \cite{ames2019controlbarrierfunction} and an \emph{HJ} \cite{bansal2017hamilton} baseline that implements a unicycle approximation for a quadruped.
For our method, we compare three different optimizers that solve the same optimization problem Eq.~\eqref{eq:opt_obj}: \emph{MPPI} with a fixed temperature, \emph{CEM}, and \emph{Replica Exchange MPPI}.
Further, while our sampling-based optimizer incurs higher per-cycle computation than CBF or HJ, this cost is already reflected in all reported metrics: the asynchronous design ensures the low-level policy continues executing at $50\, \mathrm{Hz}$ regardless of planner latency, so wall-clock optimization time affects planning staleness but never robot stability.
For completeness, we also report the performance of the unfiltered nominal contact locations plan, which serves as a lower bound on safety. 

\textbf{{Metrics:}}
We report three metrics averaged across episodes: (i)~\emph{Tracking Cost}, the cumulative tracking error between nominal and filtered contact locations plan; (ii)~\emph{Planner Violations}, the number of planning steps in which the optimized contact sequence triggers a safety violation; (iii)~\emph{Actual Violations}, the number of timesteps in which the robot makes physical contact with an obstacle.

\subsection{Simulation Results}

\begin{figure}[b]
    \centering
    \includegraphics[width=0.9\linewidth]{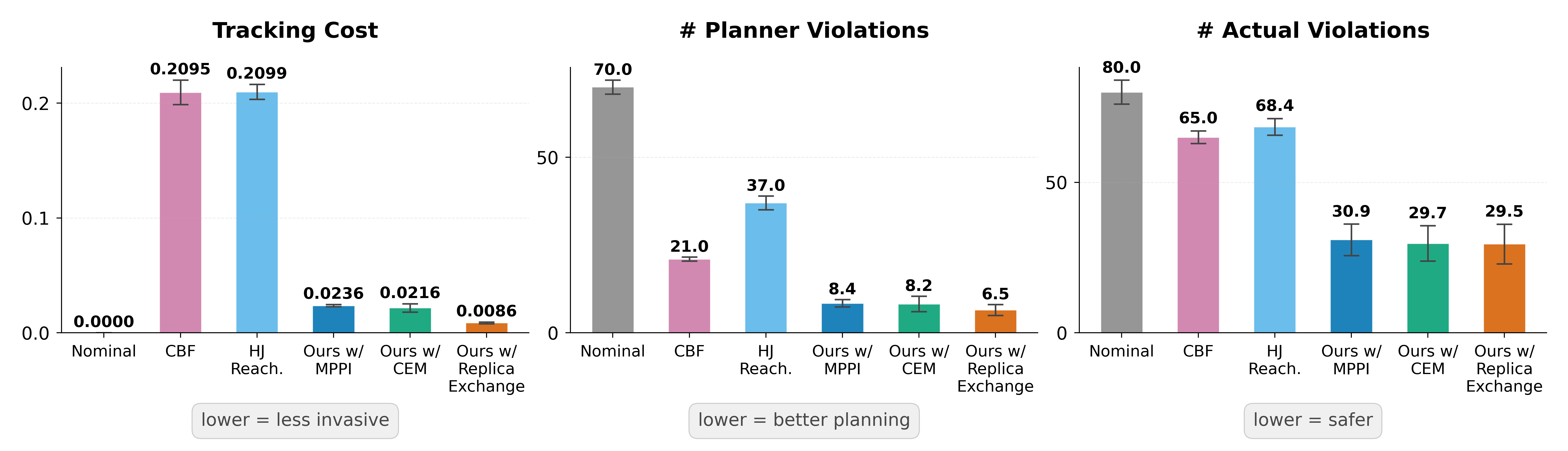}
    \vspace{-.5em}
    \caption{Comparison of tracking cost, planner violations, and actual violations across baselines and our optimizer variants. Lower is better for all metrics. Error bars show the standard error over 10 random seeds. Notably, nominal incurs zero tracking cost by definition, but serves as the lower bound on task invasiveness and upper bound on safety violations.}
    \vspace{-.5em}
    \label{fig:main_results}
\end{figure}

\begin{table}[ht]
\centering
\tiny
\setlength{\tabcolsep}{2.6pt}
\caption{Ablation on the number of iterations ($N$) and horizon length ($H$) for three optimizers with our approach. Best in each column \textbf{bolded}. Mean $\pm$ std over 10 seeds.}
\vspace{0.5em}
\label{tab:ablation_combined}
\begin{tabular}{cccccccc}
\toprule
& & \multicolumn{3}{c}{Number of Iterations ($N$)} & \multicolumn{3}{c}{Horizon Length ($H$)} \\
\cmidrule(lr){3-5} \cmidrule(lr){6-8}
\textbf{Method} & \textbf{Metric} & \textbf{1} & \textbf{3} & \textbf{6} & \textbf{4} & \textbf{6} & \textbf{8} \\
\midrule
\multirow{4}{*}{\shortstack[c]{\tiny Ours w/\\\tiny MPPI}} & Tracking Cost & $0.026\pm0.001$ & $0.024\pm0.002$ & $0.024\pm0.001$ & $0.024\pm0.001$ & $0.028\pm0.003$ & $0.031\pm0.004$ \\
 & \# Planner Viol. & $13.5\pm2.2$ & $\mathbf{8.1}\pm\mathbf{1.4}$ & $8.4\pm1.1$ & $8.4\pm1.1$ & $9.3\pm1.9$ & $10.0\pm1.5$ \\
 & \# Actual Viol. & $30.9\pm5.2$ & $33.8\pm6.1$ & $50.8\pm5.8$ & $\mathbf{30.9}\pm\mathbf{5.2}$ & $33.8\pm6.1$ & $42.5\pm4.0$ \\
 & Opt. Time (ms) & $110.6\pm4.7$ & $220.0\pm15.1$ & $570.1\pm37.5$ & $210.6\pm14.7$ & $304.2\pm25.3$ & $\mathbf{604.9}\pm\mathbf{36.2}$ \\
\cmidrule{2-8}
\multirow{4}{*}{\shortstack[c]{\tiny Ours w/\\\tiny CEM}} & Tracking Cost & $0.039\pm0.001$ & $0.021\pm0.003$ & $0.022\pm0.004$ & $0.022\pm0.004$ & $0.024\pm0.002$ & $0.020\pm0.003$ \\
 & \# Planner Viol. & $16.3\pm1.7$ & $9.7\pm1.4$ & $8.2\pm2.2$ & $8.2\pm2.2$ & $8.5\pm1.4$ & $8.2\pm1.1$ \\
 & \# Actual Viol. & $31.9\pm4.2$ & $29.7\pm5.9$ & $49.2\pm8.0$ & $31.9\pm4.2$ & $29.7\pm5.9$ & $42.7\pm5.4$ \\
 & Opt. Time (ms) & $\mathbf{107.4}\pm\mathbf{4.1}$ & $\mathbf{200.7}\pm\mathbf{16.1}$ & $\mathbf{560.7}\pm\mathbf{36.7}$ & $\mathbf{190.4}\pm\mathbf{14.1}$ & $\mathbf{263.7}\pm\mathbf{25.5}$ & $614.6\pm36.7$ \\
\cmidrule{2-8}
\multirow{4}{*}{\shortstack[c]{\tiny{Ours w/}\\\tiny{Replica}\\\tiny{Exchange}}} & Tracking Cost & $\mathbf{0.024}\pm\mathbf{0.001}$ & $\mathbf{0.010}\pm\mathbf{0.001}$ & $\mathbf{0.009}\pm\mathbf{0.001}$ & $\mathbf{0.009}\pm\mathbf{0.001}$ & $\mathbf{0.008}\pm\mathbf{0.001}$ & $\mathbf{0.006}\pm\mathbf{0.001}$ \\
 & \# Planner Viol. & $\mathbf{12.8}\pm\mathbf{2.5}$ & $9.1\pm2.0$ & $\mathbf{6.5}\pm\mathbf{1.6}$ & $\mathbf{6.5}\pm\mathbf{1.6}$ & $\mathbf{7.6}\pm\mathbf{1.3}$ & $\mathbf{7.9}\pm\mathbf{2.0}$ \\
 & \# Actual Viol. & $\mathbf{29.5}\pm\mathbf{6.6}$ & $\mathbf{28.6}\pm\mathbf{7.1}$ & $\mathbf{45.8}\pm\mathbf{4.9}$ & $31.0\pm5.0$ & $\mathbf{29.5}\pm\mathbf{6.6}$ & $\mathbf{40.2}\pm\mathbf{5.1}$ \\
 & Opt. Time (ms) & $130.2\pm5.3$ & $258.7\pm10.9$ & $638.7\pm30.8$ & $230.2\pm15.3$ & $320.9\pm28.2$ & $797.2\pm36.8$ \\
\bottomrule
\end{tabular}
\vspace{-.4em}
\end{table}

Figure~\ref{fig:main_results} shows the simulation results across three key metrics. 
The nominal plan has no tracking cost by definition, but incurs many violations since it lacks a mechanism to enforce safety. 
Both CBF and HJ Reachability reduce planner violations substantially compared to the nominal case. 
However, compared to our method, both incur higher tracking costs and more actual violations, suggesting that the reduced-order approximation is not suitable for dense, cluttered environments.

Our method, evaluated with any of the sampling-based optimizers, achieves the best safety-tracking trade-off. 
All three variants reduce actual violations to $\sim 30$ while maintaining competitive tracking costs. 
Notably, ours with Replica Exchange achieves the lowest tracking cost ($0.0086$), whereas MPPI and CEM perform similarly on all metrics. 
These results demonstrate that optimizing directly over foot contact locations with a full-physics model, rather than reduced-order dynamics, more effectively enables collision avoidance in cluttered scenes with minimal deviation from the nominal plan.
However, we acknowledge that there remains a gap between planned and actual violations.
The majority of these actual violations arise from swing-leg trajectories and the RL policy taking intermediate steps rather than unsafe contact locations.

We also ablate two key parameters of the sampling-based optimizer: the number of iterations $N$ and the planning horizon $H$, with results reported in Table~\ref{tab:ablation_combined}.
Increasing the computational budget, whether through more iterations or longer horizons, often increases actual collisions, as motion plans become stale in the asynchronous setting.
This indicates a trade-off we believe is fundamental to asynchronous predictive safety filters.
A moderate budget ($N=3$, $H \in [4 \ldots 6]$) provides a good balance.
All three optimizers behave similarly across most metrics, with Replica Exchange occasionally achieving lower tracking costs at the expense of substantially higher optimization time.
We report an additional ablation on the number of samples $K$ in the Appendix \ref{app:ablations}.

\begin{figure}[ht]
    \centering
    \includegraphics[width=0.85\linewidth]{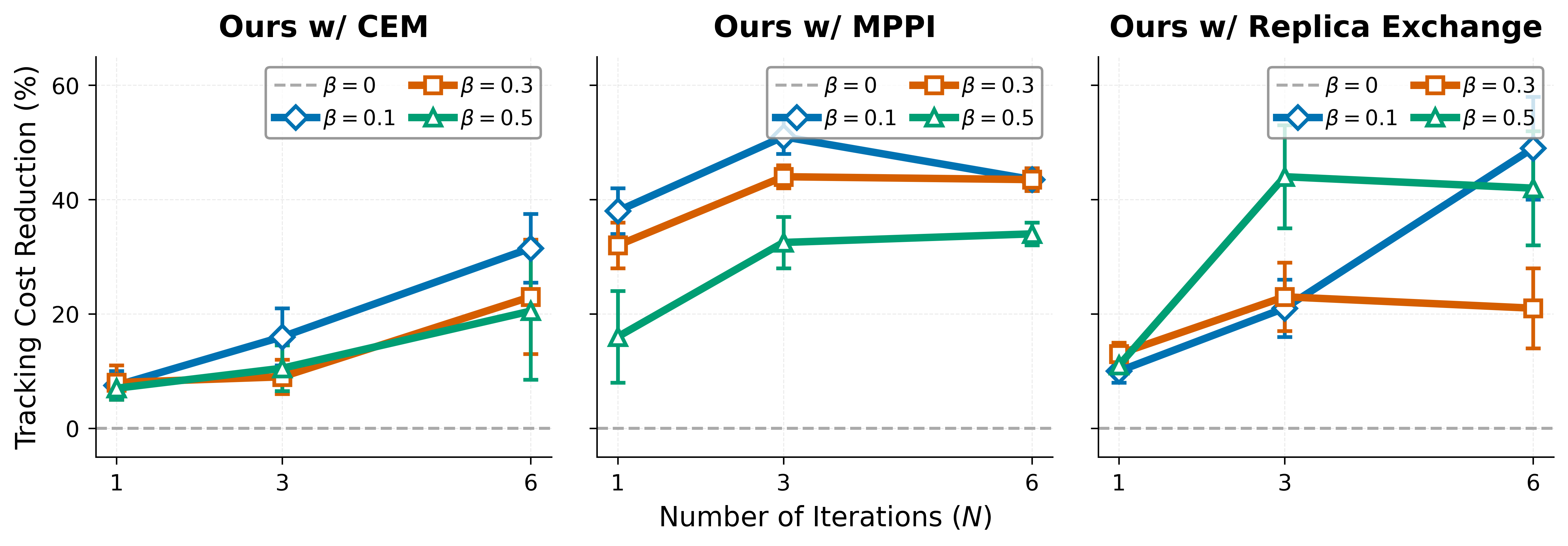}
    \vspace{-.4em}
    \caption{Percentage improvement in tracking cost relative to the baseline ($\beta = 0$, dashed), evaluated across the three optimizers at $N=\{1, 3, 6\}$ iterations. All runs are evaluated over 10 seeds.}
    \vspace{-.2em}
    \label{fig:momentum_ablation}
\end{figure}

To understand the contribution of the momentum-augmented update (Sec.~\ref{sec:mppi}), we analyze how the tracking cost improves as a function of the number of optimizer iterations $N$ and the momentum coefficient $\beta$.
Figure~\ref{fig:momentum_ablation} reports the percentage improvement in tracking cost relative to the no-momentum baseline ($\beta = 0$, dashed).
Across all three optimizers, non-zero $\beta$ yields consistent gains, with $\beta = 0.1$ achieving the best or near-best performance in most configurations.
MPPI benefits most at low iteration counts, reaching ${\sim}40\%$ improvement at a single iteration, while CEM and Replica Exchange show larger gains as $N$ increases.
Higher values ($\beta = 0.5$) exhibit greater variance, particularly in Replica Exchange, suggesting a trade-off between momentum and stability.

\subsection{Real World Validation}
\label{sec:hardware}

We deploy the predictive safety filter on a Unitree Go2 hardware, where the proprioceptive state is streamed to an external PC running the optimizer, while the computed contact targets are returned to the robot via a tethered connection. 

\begin{figure}[ht]
    \centering
    \includegraphics[width=\textwidth]{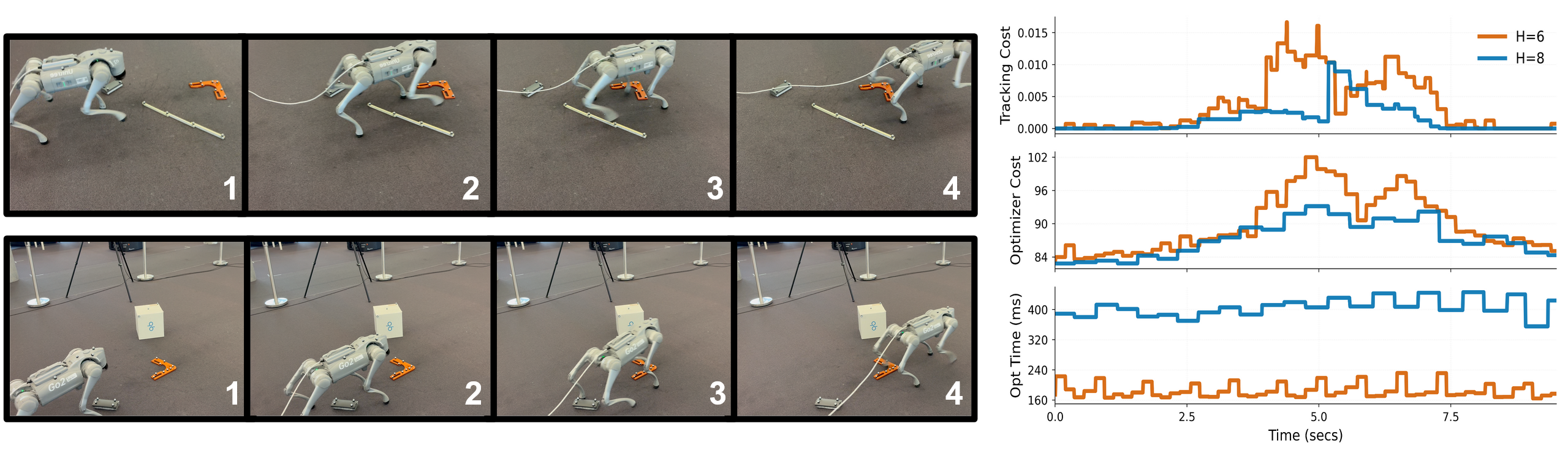}
    \vspace{-1em}
    \caption{Hardware validation on a Unitree Go2 navigating cluttered environments. \textit{Left:} Keyframes from the two scenarios. \textit{Right:} Online asynchronous planning statistics comparing $H=6$ and $H=8$. All hardware runs completed successfully.}
    \vspace{-.5em}
    \label{fig:hardware}
\end{figure}

Figure~\ref{fig:hardware} (left) shows keyframes from both the dense and the large obstacle scenarios. 
In the top sequence, the nominal contact plan would place a foot directly on an object; the filter redirects the contact to a safe gap between the objects. 
In the bottom sequence, the robot navigates around a white box and a set of poles; the filter steers the body laterally by adjusting foot placements.

Figure~\ref{fig:hardware} (right) plots the online statistics for two horizon lengths during representative trials. 
With $H=8$, the filter looks farther ahead and therefore intervenes less aggressively, resulting in lower tracking and optimizer costs. 
The price is optimization time: $H=8$ requires $\sim$$400$\,ms per cycle, whereas $H=6$ finishes in $\sim$$200$\,ms. Because the policy tracks contact targets at $50\,\mathrm {Hz}$, the robot remains stable even when the asynchronous optimizer is slow. 
This empirically validates the decoupled design: the filter can afford longer horizons when computation permits, or fall back to shorter horizons for faster reactivity, without jeopardizing low-level stability.

\section{Conclusion}

This work presents a sampling-based predictive safety filter that operates directly in the contact-location space of a contact-conditioned RL policy.
The filter optimizes foot contact targets through full-physics rollouts, bootstrapped by a learned value function and accelerated by geometric projection, momentum, and replica-exchange exploration. 
We validate the approach in dense, cluttered simulation environments and in physical hardware experiments, demonstrating a substantial reduction in safety violations while remaining minimally invasive to the nominal gait.

\textbf{Limitations \& Future Work:}
Despite strong empirical results, our method lacks theoretical guarantees of optimality or safety. 
The filtered contact plan can deviate from the true safe set in regions where the sampling-based optimizer converges to a local optimum.
Furthermore, the current geometric projection is a heuristic and may not preserve the optimizer's convergence properties. 

Actual violations in our experiments were due to the policy generating intermediate swing-stance phase configurations that violate safety despite safe contact targets.
To mitigate this, future work could implement safe online learning or fine-tuning using data collected by the optimizer.

The framework is described in general terms for any contact-conditioned legged robot, yet our validation is thus far limited to a quadruped on flat terrains. Multi-terrain and humanoid loco-manipulation introduce additional complexities in whole-body coordination and stability that remain to be tested, and scenarios requiring active torso twisting cannot be captured by contact-location optimization alone. 
Finally, integrating an online perception pipeline to enable fully autonomous deployment in unseen environments is an important direction for future work.


\clearpage
\acknowledgments{
This work was partially supported by the Huawei-TUM joint laboratory and funded in part by SIEMENS AG together with the Technical University of Munich - Institute for Advanced Study, Germany. Additional funding was provided through the Kotak IISc AI-ML Centre (KIAC) - Google Grant.
}


\bibliography{references}

\newpage
\appendix

\section{Safety Filter Algorithm}
\label{app:algo}

\begin{algorithm}[htp!]
\caption{Predictive safety filter with contact optimization}
\label{alg:filter}
\begin{algorithmic}[1]
\Require Current state $s_t$, nominal contact sequence $\{\bar{p}_{t+\tau}\}_{\tau=0}^{H-1}$, policy $\pi$, value function $V$, dynamics $f$, safety penalty parameters, optimizer hyperparameters ($K$, $N$, $\Sigma_{\text{init}}$, $\lambda_{1:L}$, $\beta$, $\alpha$, $\varepsilon_{\text{safe}}$).
\Ensure Safe contact command $p_t^*$.

\If{no safety violation predicted in nominal rollout from $s_t$}
    \Return $\bar{p}_t$
\EndIf
\If{previous optimized sequence exists}
    \State Shift it forward one step, append $\bar{p}_{t+H-1}$, set as $\mu^{\text{init}}$
\Else
    \State $\mu^{\text{init}} \gets \{\bar{p}_{t+\tau}\}_{\tau=0}^{H-1}$
\EndIf

\State Initialise $L$ replicas: mean $\mu_\ell \gets \mu^{\text{init}}$, covariance $\Sigma_\ell \gets \Sigma_{\text{init}}$, and inverse temperature $\lambda_\ell$, for $\ell=1,\dots,L$.
\For{$\text{iter}=1$ to $N$}
    \For{each replica $\ell$}
        \State Sample $K$ perturbations $\xi^{(k)} \sim \mathcal{N}(0, \Sigma_\ell)$ and form candidates $\mu_\ell^{(k)} = \mu_\ell + \xi^{(k)}$.
        \State Project all foot locations in each $\mu_\ell^{(k)}$ onto $\mathcal{F}$ via \eqref{eq:foot_proj_qp}, yielding $\tilde{\mu}_\ell^{(k)}$.
        \State Roll out all $K$ projected candidates in parallel; compute total returns $\mathcal{S}^{(k)}$ from \eqref{eq:total_obj}.
        \State $w_k \gets \exp \big(\lambda_\ell(\mathcal{S}^{(k)} - \max_j \mathcal{S}^{(j)})\big)$.
        \State Update mean: $\mu_\ell' \gets \frac{\sum_{k} w_k \, \tilde{\mu}_\ell^{(k)}}{\sum_{k} w_k}$.
        \State Apply momentum to mean: $\mu_\ell \gets \mu_\ell' + \frac{\beta}{1-\beta}(\mu_\ell' - \mu_\ell)$.
        \State Compute: $S_\ell \gets \frac{\sum_{k} w_k (\tilde{\mu}_\ell^{(k)} - \mu_\ell)(\tilde{\mu}_\ell^{(k)} - \mu_\ell)^\top}{\sum_{k} w_k}$.
        \State Update covariance: $\Sigma_\ell \gets S_\ell + \frac{\beta(\Sigma_\ell - S_\ell) + 2\alpha S_\ell}{\beta + 1 - 2\alpha}$.
    \EndFor
    \State For adjacent replicas $i, j$: propose to swap $(\mu_i,\Sigma_i)$ and $(\mu_j,\Sigma_j)$ with probability from \eqref{eq:replica_swap}, using $\bar{\mathcal{S}}_i, \bar{\mathcal{S}}_j$.
\EndFor
\State Project the optimized sequence $\mu_1$ onto $\mathcal{F}$ via Eq.~\eqref{eq:foot_proj_qp}, yielding final feasible plan $\tilde{\mu}_1$.
\State \Return first timestep contact location of $\tilde{\mu}_1$ as $p_t^*$.
\end{algorithmic}
\end{algorithm}

\section{Derivation of Sampling-Based Optimization Updates}
\label{app:derivation}

This appendix derives the momentum-augmented mean and covariance updates (Equations~\eqref{eq:mu_update}--\eqref{eq:sigma_update}) used in Section~\ref{sec:mppi}.
We work throughout in the \emph{reward-maximization} setting of the main paper, using the total return $\mathcal{S}$ defined in Equation~\eqref{eq:total_obj}, so no sign change is required.
Section~\ref{app:fkl_fixed_points} shows how both CEM and MPPI emerge from a natural-gradient projection of a reward-weighted target distribution onto the Gaussian family.
Section~\ref{app:accel_entropy} then derives the entropy-regularized, momentum-augmented extension used in our method.

\subsection{Forward-KL Gaussian Fitting and Moment-Matching Fixed Points}
\label{app:fkl_fixed_points}

Let $p_\theta(x) = \mathcal{N}(x \mid \mu, \Sigma)$ be the Gaussian search distribution with parameters $\theta = (\mu, \Sigma)$, where $\mu \in \mathbb{R}^d$ and $\Sigma \in \mathbb{S}^d_{++}$ (symmetric positive definite).

At optimizer iteration $t$, with current parameters $\theta_t = (\mu_t, \Sigma_t)$ and inverse temperature $\lambda > 0$, we define the \emph{reward-weighted} target
\begin{equation}
    q_t(x) \;:=\; \frac{p_{\theta_t}(x)\,\exp \bigl(\lambda\,\mathcal{S}(x)\bigr)}{Z_t},
    \qquad
    Z_t \;:=\;\int p_{\theta_t}(x)\,\exp \bigl(\lambda\,\mathcal{S}(x)\bigr)\,\mathrm{d}x.
    \label{eq:app_target}
\end{equation}
High-return regions of the search space are up-weighted according to $\lambda$; as $\lambda \to \infty$ the target concentrates on the global maximizer of $\mathcal{S}$.

We obtain the next search distribution by projecting $q_t$ onto the Gaussian family via the forward KL divergence:
\begin{equation}
    \theta_{t+1}
    \;=\; \argmin_\theta\;
    D_{\mathrm{KL}}\!\bigl(q_t \;\big\|\; p_\theta\bigr).
    \label{eq:app_fkl}
\end{equation}
Since $q_t$ is fixed with respect to $\theta$, minimizing the forward KL is equivalent to maximum-likelihood fitting of $p_\theta$ under $q_t$ \cite{de2005tutorial}:
\begin{equation}
    \theta_{t+1} \;=\; \argmax_\theta\; \mathbb{E}_{x \sim p_{\theta_t}}\!\bigl[w(x)\,\log p_\theta(x)\bigr],
    \label{eq:app_mle}
\end{equation}
where the expectation has been converted from $q_t$ to $p_{\theta_t}$ via \emph{self-normalized importance weights}:
\begin{equation}
    w(x) \;:=\; \frac{q_t(x)}{p_{\theta_t}(x)}
    \;=\; \frac{\exp \bigl(\lambda\,\mathcal{S}(x)\bigr)}{Z_t}
    \;\approx\;
    \frac{\exp \bigl(\lambda\,\mathcal{S}(x)\bigr)} {\mathbb{E}_{x'\sim p_{\theta_t}}\!\bigl[\exp \bigl(\lambda\,\mathcal{S}(x')\bigr)\bigr]},
    \label{eq:app_weights}
\end{equation}
where the denominator is approximated by a sample average (self-normalized importance sampling). 
By construction, $w(x) \ge 0$ and $\mathbb{E}_{x \sim p_{\theta_t}}[w(x)] = 1$.

Let $\mathcal{J}_{\mathrm{KL}}(\theta) := D_{\mathrm{KL}}(q_t \| p_\theta)$.
Using the proximal natural-gradient formulation of \cite{wagener2019online}, the solution of~\eqref{eq:app_fkl} can be computed via the natural gradient descent step $\theta_{t+1} = \theta_t - \eta\, F(\theta_t)^{-1}\nabla_\theta \mathcal{J}_{\mathrm{KL}}(\theta_t)$, where $F(\theta)$ is the Fisher information matrix of $p_\theta$ and $\eta > 0$ is the step size.
For the mean-covariance parameterization of a Gaussian, the natural gradients evaluate to \cite{de2005tutorial}:
\begin{align}
   F(\theta)^{-1}\nabla_\mu\,\mathcal{J}_{\mathrm{KL}}(\mu, \Sigma) &\;=\; \mu \;-\; \mathbb{E}_{x \sim p_{\theta_t}}\!\bigl[w(x)\,x\bigr],
   \label{eq:app_ng_mu} \\[4pt]
   F(\theta)^{-1}\nabla_\Sigma\,\mathcal{J}_{\mathrm{KL}}(\mu, \Sigma) &\;=\; \mathbb{E}_{x \sim p_{\theta_t}}\! \bigl[w(x)\,(x - \mu)(x - \mu)^\top\bigr] \;-\; \Sigma.
   \label{eq:app_ng_sigma}
\end{align}
Setting both to zero yields the \emph{reward-weighted moment-matching fixed points}:
\begin{align}
   \mu^{\star} &\;=\; \underbrace{%
            \mathbb{E}_{x \sim p_{\theta_t}}\!\bigl[w(x)\,x\bigr]
            }_{\displaystyle:=\; \mu_t^\prime},
    \label{eq:app_fp_mu} \\[6pt]
   \Sigma^{\star} &\;=\; \underbrace{%
            \mathbb{E}_{x \sim p_{\theta_t}}\!
            \bigl[w(x)\,(x - \mu_t^\prime)(x - \mu_t^\prime)^\top\bigr]
            }_{\displaystyle:=\; S(\mu_t^\prime)}.
     \label{eq:app_fp_sigma}
\end{align}

A key property is that the Eq.~\eqref{eq:app_fkl} depends on $w(x)$ only through its expectation, so the fixed points~\eqref{eq:app_fp_mu}--\eqref{eq:app_fp_sigma} hold for \emph{any} non-negative unit-mean weighting scheme.
Two standard choices for this used in our method are:

\noindent\textbf{MPPI}\cite{williams2015modelpredictivepathintegral} uses \emph{soft} exponential weights over $K$ samples, numerically stabilized by subtracting the batch maximum:
\begin{equation}
   w_k \;=\; \frac{\exp \bigl(\lambda\,(\mathcal{S}^{(k)} - \max_i \mathcal{S}^{(i)})\bigr)}
        {\displaystyle\sum_{k=1}^K
        \exp \bigl(\lambda\,(\mathcal{S}^{(k)} - \max_i \mathcal{S}^{(i)})\bigr)}.
   \label{eq:app_mppi_w}
\end{equation}

\noindent\textbf{CEM} \cite{rubinstein1999cross} uses \emph{hard elite} weights: let $\mathcal{S}_{(K_e)}$ denote the $K_e$-th largest return among $K$ samples. Then
\begin{equation}
   w_k
   \;=\;
   \frac{K}{K_e}\,\mathbb{I}\!\bigl[\mathcal{S}^{(k)} \ge \mathcal{S}_{(K_e)}\bigr],
   \label{eq:app_cem_w}
\end{equation}
satisfying $\frac{1}{K}\sum_k w_k = 1$.
Substituting into~\eqref{eq:app_fp_mu}--\eqref{eq:app_fp_sigma} recovers the standard CEM update (sample mean and covariance of the elite set).

\subsection{Accelerated Entropy-Regularized Updates}
\label{app:accel_entropy}

We now extend the Gaussian fitting framework of Section~\ref{app:fkl_fixed_points} to incorporate (i) an entropy bonus that prevents premature collapse of the search distribution and (ii) Nesterov-style momentum that accelerates convergence.
These extensions yield Eqs.~\eqref{eq:mu_update}--\eqref{eq:sigma_update}.

Augmenting the forward-KL objective with the differential entropy of $p_\theta$ yields:
\begin{equation}
   \mathcal{J}(\theta) \;:= D_{\mathrm{KL}}\!\bigl(q_t \;\big\|\; p_\theta\bigr) \;-\; \alpha\,\mathcal{H}(p_\theta), \qquad \alpha \;\ge\; 0,
   \label{eq:app_reg_obj}
\end{equation}
where for $\alpha > 0$, minimizing $\mathcal{J}$ simultaneously aligns $p_\theta$ with $q_t$ and maximizes its entropy, keeping the search distribution broad.

\paragraph{Accelerated proximal natural-gradient objective.}
Momentum is incorporated following the Bayesian learning rule
of \cite{khan2023bayesian}, which augments the proximal objective with a \emph{repulsion} term from the previous iterate $\theta_{t-1}$:
\begin{equation}
   \theta_{t+1} \;=\; \argmin_\theta\;
   \Bigl[
     \bigl\langle\nabla_\theta\mathcal{J}(\theta_t),\;\theta - \theta_t\bigr\rangle 
     + \frac{1 + \beta}{\eta}\, D_{\mathrm{KL}}\!\bigl(p_\theta \;\big\|\; p_{\theta_t}\bigr)
     - \frac{\beta}{\eta}\, D_{\mathrm{KL}}\!\bigl(p_\theta \;\big\|\; p_{\theta_{t-1}}\bigr)
   \Bigr],
   \label{eq:app_accel_prox}
\end{equation}
where $\eta > 0$ is the step size and $\beta \ge 0$ is the momentum coefficient.
The $\frac{1+\beta}{\eta}$ term anchors $\theta_{t+1}$ to the current iterate $\theta_t$, while the $-\frac{\beta}{\eta}$ term repels it from the previous iterate $\theta_{t-1}$, producing an extrapolation in the direction of improvement analogous to Nesterov acceleration.

Setting the first-order variations of~\eqref{eq:app_accel_prox} with respect to $\mu_{t+1}$ and $\Sigma_{t+1}$ to zero, using the natural-gradient expressions derived in \cite{khan2023bayesian} for the entropy-regularized Gaussian objective~\eqref{eq:app_reg_obj}, gives:
\begin{align}
   0 &\;=\; -\eta\,(\mu_{t+1} - \mu_t^\prime) \;+\; \beta\,(\mu_{t+1} - \mu_t),
    \label{eq:app_stat_mu} \\[6pt]
   0 &\;=\; -\eta\,\bigl[S(\mu_{t+1}) - (1 - 2\alpha)\,\Sigma_{t+1}\bigr] \;+\; \beta\,(\Sigma_{t+1} - \Sigma_t),
    \label{eq:app_stat_sigma}
\end{align}
where $\mu_t^\prime$ and $S(\mu)$ are defined in~\eqref{eq:app_fp_mu} and~\eqref{eq:app_fp_sigma}.
The factor $(1 - 2\alpha)$ in~\eqref{eq:app_stat_sigma} arises from the natural gradient of $-\alpha\mathcal{H}$ in the exponential-family parameterization \cite{khan2023bayesian}: the entropy bonus contributes an additional $+2\alpha\Sigma$ to the stationarity condition, enlarging the equilibrium covariance relative to the unregularized case.
Setting $\alpha = 0$ in~\eqref{eq:app_stat_sigma} recovers the stationarity condition of the pure-KL objective from Section~\ref{app:fkl_fixed_points}.

\paragraph{Practical update rules.}
Equations~\eqref{eq:app_stat_mu}--\eqref{eq:app_stat_sigma} still depend on the step size $\eta$.
To obtain step-size-free update rules, we apply the reparameterization $\beta \leftarrow \beta/\eta$, which absorbs $\eta$ into the momentum coefficient \cite{khan2023bayesian}.
Under this substitution, dividing~\eqref{eq:app_stat_mu}--\eqref{eq:app_stat_sigma} by $\eta$ and solving the resulting linear equations in closed form yields:

\noindent\textit{Mean.}
From~\eqref{eq:app_stat_mu}:
\begin{equation}
   (1 - \beta)\,\mu_{t+1} \;=\; \mu_t^\prime - \beta\,\mu_t
   \qquad\Longrightarrow\qquad
   \boxed{
   \mu_{t+1}
   \;=\; \mu_t^\prime
         \;+\; \frac{\beta}{1 - \beta}\,(\mu_t^\prime - \mu_t).
   }
   \label{eq:app_mu_update}
\end{equation}

\noindent\textit{Covariance.}
From~\eqref{eq:app_stat_sigma}:
\begin{equation}
\begin{aligned}
   (\beta + 1 - 2\alpha)\,\Sigma_{t+1}
   \;=\; S(\mu_{t+1}) &+ \beta\,\Sigma_t
   \qquad\Longrightarrow\qquad \\
   &\boxed{
   \Sigma_{t+1}
   \;=\; S(\mu_{t+1})
         \;+\; \frac{\beta\,\bigl(\Sigma_t - S(\mu_{t+1})\bigr)
                      + 2\alpha\,S(\mu_{t+1})}
                     {\beta + 1 - 2\alpha}.
   }
   \label{eq:app_sigma_update}    
\end{aligned}
\end{equation}
Note that the second moment $S(\mu_{t+1})$ is centered at the \emph{updated} mean $\mu_{t+1}$ (Eq.~\eqref{eq:app_mu_update}), not at $\mu_t$; in practice, $S(\mu_{t+1})$ is computed after the mean update. The two boxed equations correspond exactly to Eqs.~\eqref{eq:mu_update} and~\eqref{eq:sigma_update} in Sec. \ref{sec:mppi}.

The mean update~\eqref{eq:app_mu_update} extrapolates beyond the current reward-weighted centroid $\mu_t^\prime$ by a fraction $\beta/(1-\beta)$ in the direction of improvement
$\mu_t^\prime - \mu_t$.
Setting $\beta = 0$ recovers the vanilla fixed point $\mu_{t+1} = \mu_t^\prime$.

The covariance update~\eqref{eq:app_sigma_update} can be written equivalently as,
\begin{equation}
   \Sigma_{t+1} \;=\; \frac{S(\mu_{t+1}) + \beta\,\Sigma_t}{\beta + 1 - 2\alpha},
   \label{eq:app_sigma_alt}
\end{equation}
which reveals it as a weighted blend of the current $S(\mu_{t+1})$ and the previous covariance $\Sigma_t$, divided by $(\beta + 1 - 2\alpha)$.
For $\alpha > 0$, this denominator is less than $(\beta + 1)$, inflating the resulting covariance to maintain exploration.
Setting $\beta = \alpha = 0$ recovers the vanilla fixed point $\Sigma_{t+1} = S(\mu_t^\prime)$.

Lastly, the denominator $\beta + 1 - 2\alpha$ in~\eqref{eq:app_sigma_update}--\eqref{eq:app_sigma_alt} must be strictly positive, which requires
\begin{equation}
   0 \;\le\; \alpha \;<\; \frac{\beta + 1}{2}.
   \label{eq:app_reg_cond}
\end{equation}

\paragraph{Applicability to both MPPI and CEM.}
Updates~\eqref{eq:app_mu_update}--\eqref{eq:app_sigma_update} depend on the data only through $\mu_t^\prime$ and $S(\mu_{t+1})$, which are weighted sums parameterized by the normalized weights $\{w_k\}$.
They therefore hold verbatim for both the MPPI soft weights~\eqref{eq:app_mppi_w} and the CEM hard elite weights~\eqref{eq:app_cem_w}.
In Sec. \ref{sec:mppi} and Algorithm \ref{alg:filter}, MPPI weights are used due to the incorporation of Replica Exchange; However, we do provide an Algorithm~\ref{alg:aercem} below that also uses CEM weights for concreteness.

\begin{algorithm}[h]
\caption{Accelerated Entropy-Regularized Sampling-Based Optimizer}
\label{alg:aercem}
\begin{algorithmic}[1]
\Require
 Initial mean $\mu_0 \in \mathbb{R}^d$,
 covariance $\Sigma_0 \in \mathbb{S}^d_{++}$,
 momentum $\beta \in [0,\,1)$,
 entropy coefficient $\alpha \ge 0$ with $\alpha < (\beta{+}1)/2$,
 sample budget $K$,
 number of elite samples $K_e \le K$,
 inverse temperature $\lambda > 0$
\State Set $\mu_{-1} \leftarrow \mu_0$
\While{not converged}
 \State \textbf{Sample}\;
        $x_k \sim \mathcal{N}(\mu_t,\,\Sigma_t)$
        for $k = 1, \ldots, K$
 \State \textbf{Evaluate}\;
        total return $\mathcal{S}(x_k)$ for each $k$
 \State \textbf{Compute weights}:
   \Statex\quad
     \textit{CEM}: $w_k = \tfrac{K}{K_e}\,
       \mathbb{I}[\mathcal{S}^{(k)} \ge \mathcal{S}_{(K_e)}]$
   \Statex\quad
     \textit{MPPI}: $w_k \propto \exp \bigl(
       \lambda(\mathcal{S}^{(k)} - \max_i \mathcal{S}^{(i)})\bigr)$,
       renormalized to $\textstyle\sum_k w_k = 1$
 \State $\mu_t^\prime \;\leftarrow\; \sum_{k=1}^{K} w_k\,x_k$
 \State $\mu_{t+1} \;\leftarrow\;
        \mu_t^\prime + \dfrac{\beta}{1-\beta}(\mu_t^\prime - \mu_t)$
 \State $S_t \;\leftarrow\; \sum_{k=1}^{K}
        w_k\,(x_k - \mu_{t+1})(x_k - \mu_{t+1})^\top$
 \State $\Sigma_{t+1} \;\leftarrow\; S_t +
        \dfrac{\beta\,(\Sigma_t - S_t) + 2\alpha\,S_t}
              {\beta + 1 - 2\alpha}$
 \State $t \leftarrow t + 1$
\EndWhile
\end{algorithmic}
\end{algorithm}

\section{Additional Ablations}
\label{app:ablations}

Table~\ref{tab:ablation_projection} ablates the projection configuration in our safety filter across three optimizer variants.
Projection \emph{in} projects sampled foot placements during optimizer rollouts, while projection \emph{out} applies a final projection to the optimized contact plan before execution by the low-level RL policy.
Applying both projections consistently yields the best safety-performance tradeoff across all optimizers.

\begin{table}[ht]
\centering
\scriptsize
\setlength{\tabcolsep}{3.2pt}
\caption{Ablation on projection configuration across three optimizers.
\textbf{in} = projection applied during rollout after sampling, \textbf{out} = projection applied on the final optimized plan that is sent to the low-level RL policy. The runs are over 10 seeds with the best in each algorithm group \textbf{bolded}.}
\label{tab:ablation_projection}
\vspace{0.5em}
\begin{tabular}{lcccccc}
\toprule
& \multicolumn{2}{c}{Projection} & & & & \\
\cmidrule(lr){2-3}
\textbf{Method} & \textbf{in} & \textbf{out} & \textbf{Tracking Cost $\downarrow$} & \textbf{\# Planner Viol. $\downarrow$} & \textbf{\# Actual Viol. $\downarrow$} \\
\midrule
\multirow{4}{*}{\shortstack[c]{Ours w/\\MPPI}}
  & -- & -- & $0.028\pm0.003$ & $25.3\pm3.2$ & $51.2\pm6.1$ \\
  & -- & $\checkmark$ & $0.023\pm0.002$ & $15.1\pm2.1$ & $\mathbf{29.4}\pm\mathbf{5.3}$ \\
  & $\checkmark$ & -- & $\mathbf{0.019}\pm\mathbf{0.002}$ & $17.8\pm2.8$ & $35.8\pm5.7$ \\
  & $\checkmark$ & $\checkmark$ & $0.020\pm0.002$ & $\mathbf{9.1}\pm\mathbf{1.5}$ & $30.3\pm5.0$ \\
\midrule
\multirow{4}{*}{\shortstack[c]{Ours w/\\Replica\\Exchange}}
  & -- & -- & $0.011\pm0.001$ & $19.8\pm3.1$ & $42.1\pm6.4$ \\
  & -- & $\checkmark$ & $0.010\pm0.001$ & $13.2\pm2.2$ & $32.3\pm5.6$ \\
  & $\checkmark$ & -- & $\mathbf{0.008}\pm\mathbf{0.001}$ & $12.4\pm1.9$ & $35.6\pm5.9$ \\
  & $\checkmark$ & $\checkmark$ & $0.009\pm0.001$ & $\mathbf{7.9}\pm\mathbf{1.3}$ & $\mathbf{27.8}\pm\mathbf{5.2}$ \\
\midrule
\multirow{4}{*}{\shortstack[c]{Ours w/\\CEM}}
  & -- & -- & $0.023\pm0.003$ & $23.7\pm3.5$ & $48.6\pm6.8$ \\
  & -- & $\checkmark$ & $0.026\pm0.003$ & $15.9\pm2.4$ & $34.8\pm5.8$ \\
  & $\checkmark$ & -- & $\mathbf{0.018}\pm\mathbf{0.002}$ & $13.1\pm1.8$ & $38.7\pm6.2$ \\
  & $\checkmark$ & $\checkmark$ & $0.023\pm0.003$ & $\mathbf{9.2}\pm\mathbf{1.4}$ & $\mathbf{29.1}\pm\mathbf{5.1}$ \\
\bottomrule
\end{tabular}
\end{table}

The ablation also highlights the distinct roles of the two projection stages.
Using only the rollout-time projection (\emph{in}) improves safety during optimization by biasing samples toward feasible regions, but it does not guarantee that the final optimized trajectory remains feasible after multiple optimizer updates.
This issue is especially pronounced for sampling-based optimizers, where iterative reweighting and distribution updates can collapse the sampling distribution toward a single mode that drifts outside the feasible set.
As a result, even if intermediate samples are projected, the final optimized contact sequence may still have violations.
The final projection (\emph{out}) is therefore critical because it explicitly enforces feasibility on the executed plan after the optimizer converges.
Lastly, we found that projection was also the most significant component of our proposed algorithmic improvements (projection, momentum, and replica exchange) to reduce the number of safety violations in planning and execution.

Table~\ref{tab:ablation_value} ablates the learned terminal value function used in \eqref{eq:opt_obj}.
Across all three optimizer variants (Replica Exchange, MPPI, CEM), enabling the learned value function yields large improvements. 
Turning it on cuts tracking cost, however, marginal gains can be seen in terms of planner and actual violations.
This is attributable to the RL policy's swing-phase behavior and its poor contact-goal-reaching capability, which contact-space optimization cannot directly address.

\begin{table}[h]
\centering
\scriptsize
\setlength{\tabcolsep}{2.6pt}
\caption{Ablation on the value function component across three optimizers.
\textbf{w/o} = component disabled, \textbf{w/} = component enabled. Best in each pair \textbf{bolded}. Mean $\pm$ std over 10 seeds.}
\vspace{0.5em}
\label{tab:ablation_value}
\begin{tabular}{cccccccc}
\toprule
& & \multicolumn{2}{c}{Tracking Cost $\downarrow$} & \multicolumn{2}{c}{\# Planner Violations $\downarrow$} & \multicolumn{2}{c}{\# Actual Violations $\downarrow$} \\
\cmidrule(lr){3-4} \cmidrule(lr){5-6} \cmidrule(lr){7-8}
\textbf{Component} & \textbf{Method} & \textbf{w/o} & \textbf{w/} & \textbf{w/o} & \textbf{w/} & \textbf{w/o} & \textbf{w/} \\
\midrule
\multirow{3}{*}{\shortstack[c]{Value\\Function}}
  & Ours w/ Replica Exchange & $0.012\pm0.001$ & $\mathbf{0.011}\pm\mathbf{0.001}$ & $13.1\pm3.8$ & $\mathbf{10.2}\pm\mathbf{3.1}$ & $30.1\pm6.2$ & $\mathbf{29.1}\pm\mathbf{7.1}$ \\
  & Ours w/ MPPI & $0.029\pm0.002$ & $\mathbf{0.020}\pm\mathbf{0.001}$ & $\mathbf{9.3}\pm\mathbf{1.6}$ & ${10.4}\pm{4.3}$ & $\mathbf{30.3}\pm\mathbf{7.4}$ & ${31.6}\pm{6.2}$ \\
  & Ours w/ CEM & $0.033\pm0.001$ & $\mathbf{0.022}\pm\mathbf{0.001}$ & $11.0\pm3.4$ & $\mathbf{9.1}\pm\mathbf{1.5}$ & $33.2\pm6.3$ & $\mathbf{32.6}\pm\mathbf{5.4}$ \\
\bottomrule
\end{tabular}
\end{table}

\begin{table}[h]
\centering
\scriptsize
\setlength{\tabcolsep}{4pt}
\caption{Ablation on the number of samples ($K$) in our approach. Best in each column \textbf{bolded} with mean $\pm$ std over 10 seeds.}
\vspace{0.5em}
\label{tab:ablation_samples}
\begin{tabular}{llccc}
\toprule
& & \multicolumn{3}{c}{Number of Samples ($K$)} \\
\cmidrule(lr){3-5}
\textbf{Method} & \textbf{Metric} & \textbf{32} & \textbf{256} & \textbf{512} \\
\midrule
\multirow{4}{*}{\shortstack[c]{{Ours w/}\\{MPPI}}} & Tracking Cost & $0.046\pm0.001$ & $0.031\pm0.001$ & $0.024\pm0.001$ \\
 & \# Planner Viol. & $32.9\pm2.3$ & $15.7\pm0.9$ & $8.4\pm1.1$ \\
 & \# Actual Viol. & $\mathbf{42.5}\pm\mathbf{4.0}$ & $\mathbf{30.9}\pm\mathbf{5.2}$ & $34.1\pm4.5$ \\
 & Opt. Time (ms) & $110.6\pm4.7$ & $\mathbf{143.3}\pm\mathbf{4.5}$ & $171.0\pm4.8$ \\
\cmidrule{2-5}
\multirow{4}{*}{\shortstack[c]{{Ours w/}\\{CEM}}} & Tracking Cost & $0.049\pm0.001$ & $0.031\pm0.002$ & $0.022\pm0.004$ \\
 & \# Planner Viol. & $34.6\pm1.9$ & $18.0\pm1.4$ & $8.2\pm2.2$ \\
 & \# Actual Viol. & $49.7\pm5.9$ & $31.9\pm4.2$ & $32.7\pm5.4$ \\
 & Opt. Time (ms) & $\mathbf{107.4}\pm\mathbf{4.1}$ & $148.7\pm5.6$ & $\mathbf{166.6}\pm\mathbf{3.9}$ \\
\cmidrule{2-5}
\multirow{4}{*}{\shortstack[c]{{Ours w/}\\{Replica}\\{Exchange}}} & Tracking Cost & $\mathbf{0.044}\pm\mathbf{0.001}$ & $\mathbf{0.014}\pm\mathbf{0.002}$ & $\mathbf{0.009}\pm\mathbf{0.001}$ \\
 & \# Planner Viol. & $\mathbf{30.8}\pm\mathbf{1.5}$ & $\mathbf{14.1}\pm\mathbf{1.9}$ & $\mathbf{6.5}\pm\mathbf{1.6}$ \\
 & \# Actual Viol. & $43.8\pm4.3$ & $31.0\pm5.0$ & $\mathbf{29.5}\pm\mathbf{6.6}$ \\
 & Opt. Time (ms) & $130.2\pm5.3$ & $190.1\pm14.6$ & $211.4\pm12.5$ \\
\bottomrule
\end{tabular}
\end{table}

Lastly, Table~\ref{tab:ablation_samples} shows how the sample budget $K$ affects performance and computational cost.
For all methods, increasing $K$ from 32 to 512 yields a large improvement in tracking cost and planner violations, with diminishing returns beyond $K=256$.  
Interestingly, actual safety violations do not decrease monotonically: the lowest actual violation counts are achieved at $K=256$ for MPPI and CEM, and at $K=512$ for Replica Exchange.  
This suggests that, with a large number of samples, the planner does not necessarily reduce collisions further.
Replica Exchange consistently outperforms the single-temperature baselines across all sample sizes, and its actual violation continues to improve at $K=512$ thanks to the parallel-tempering exploration. 
Optimization time grows roughly linearly with $K$; Replica Exchange adds a 30-60\% overhead relative to MPPI/CEM at the same sample count due to the multiple replicas, but its superior performance often justifies this cost.

\section{Implementation Details}

\subsection{RL Policy Architecture}
\label{app:policy}

The contact-conditioned locomotion policy is a pre-trained actor network (trained offline similarly to \cite{omar2025learningactcontactunified}) and is frozen during deployment.  
It receives an 82-dimensional observation vector and outputs 12-dimensional joint position targets:

\begin{itemize}[noitemsep, topsep=0pt, partopsep=0pt]
    \item \textit{Architecture}: 4-layer MLP (512 $\to$ 256 $\to$ 128 $\to$ 12) with ELU activations.
    \item \textit{Observation} (82 dims): base linear velocity (3), base angular velocity (3), projected gravity (3), planned contact locations in base frame for the next two timesteps (24 = 4 feet $\times$ 2 references $\times$ 3D), time remaining in the current gait phase (1), current swing/stance flags (8 = 4 feet $\times$ 2 phases), joint positions relative to default (12), joint velocities (12), foot end-effector position error norms (4 = 4 feet $\times$ L2 norm of error), last action (12).
    \item \textit{Action}: Joint position targets offset from the nominal stance, scaled by a factor of 0.35 to limit deviation.
    \item \textit{PD Control}: Joint torques are computed with $k_p = 30.0$, $k_d = 0.5$, and joint position limits are softly enforced by scaling down commands when the joint is within 5\% of its limit (95\% soft factor).
\end{itemize}

\subsection{Geometric Projection}
\label{app:projection}

All obstacles are modeled as convex geometric primitives on the ground plane ($xy$-plane), and their SDFs and gradients admit closed-form expressions,  summarized below. 
In all cases, $\mathbf{R}_j \in SO(2)$ denotes the primitive's orientation and $\mathbf{q} \triangleq \mathbf{R}_j^\top(\mathbf{p} - \mathbf{c}_j)$ its expression in the local frame.

\textit{Circle / Cylinder.}
\begin{equation}
    \mathcal{D}_j(\mathbf{p}) = \|\mathbf{p} - \mathbf{c}_j\| - r_j, \qquad \nabla\mathcal{D}_j(\mathbf{p}) = \frac{\mathbf{p} - \mathbf{c}_j}{\|\mathbf{p} - \mathbf{c}_j\|}.
\end{equation}

\textit{Box.} Let $\mathbf{e}_j = [e_x,\, e_y]^\top$ be the half-extents and  $\mathbf{d} = |\mathbf{q}| - \mathbf{e}_j$ the signed per-axis residual. 
Then
\begin{equation}
    \mathcal{D}_j(\mathbf{p}) = \underbrace{\|\max(\mathbf{d},\,\mathbf{0})\|}_{\text{exterior}} + \underbrace{\min\!\bigl(\max(d_x,\, d_y),\, 0\bigr)}_{\text{interior}}.
\end{equation}
The gradient in the local frame is
\begin{equation}
    \nabla_{\mathbf{q}}\mathcal{D}_j =
    \begin{cases}
        \dfrac{\operatorname{sign}(\mathbf{q}) \odot \max(\mathbf{d},\mathbf{0})}{\|\max(\mathbf{d},\mathbf{0})\|} & \text{if } \mathbf{p} \notin \mathcal{B}_j, \\[6pt]
        \operatorname{sign}(q_i)\,\mathbf{e}_i, \quad i = \arg\max_k\, d_k & \text{if } \mathbf{p} \in \mathcal{B}_j,
    \end{cases}
\end{equation}
where $\mathcal{B}_j$ denotes the box interior and $\mathbf{e}_i$ is the $i$-th standard basis vector. 
The world-frame gradient follows as $\nabla_{\mathbf{p}}\mathcal{D}_j = \mathbf{R}_j\,\nabla_{\mathbf{q}}\mathcal{D}_j$.

\textit{Capsule.} Let $l_j$ be the half-length along the local $x$-axis. 
Define the nearest point on the capsule axis as $\mathbf{q}_{\text{proj}} = [\operatorname{clip}(q_x,\,-l_j,\,l_j),\; 0]^\top$, and let $\boldsymbol{\delta} = \mathbf{q} - \mathbf{q}_{\text{proj}}$. Then
\begin{equation}
    \mathcal{D}_j(\mathbf{p}) = \|\boldsymbol{\delta}\| - r_j, \qquad \nabla_{\mathbf{p}}\mathcal{D}_j(\mathbf{p}) = \mathbf{R}_j\, \frac{\boldsymbol{\delta}}{\|\boldsymbol{\delta}\|}.
\end{equation}

These SDFs and gradients are evaluated in a single batched call over all obstacles and all $K$ sampled foot positions, with no branching overhead at runtime. 
To project a candidate foot contact location $p_t^{(k)}$ onto $\mathcal{F}$, we then linearize each SDF constraint via a first-order Taylor expansion,
\begin{equation}
    \nabla\mathcal{D}_j\!\left(p_t^{(k)}\right)^\top\!\!\left(\mathbf{p} - p_t^{(k)}\right) \geq \varepsilon_{\text{safe}} - \mathcal{D}_j\!\left(p_t^{(k)}\right),
\end{equation}
reducing the projection to the QP
\begin{equation}
    \min_{\mathbf{p}} \; \|\mathbf{p} - p_t^{(k)}\|^2 \quad \text{s.t.} \quad \mathbf{a}_j^\top \mathbf{p} \leq b_j \;\; \forall\, j,
    \label{eq:foot_proj_qp}
\end{equation}
where $\mathbf{a}_j = -\nabla\mathcal{D}_j(p_t^{(k)})$ and $b_j = \varepsilon_{\text{safe}} - \mathcal{D}_j(p_t^{(k)}) - \mathbf{a}_j^\top p_t^{(k)}$. 
This small QP can then be solved using an active-set method or an off-the-shelf solver.

\subsection{Value Function Training}
\label{app:value_training}

The terminal value $V(s)$ approximates the expected sum of future returns from state $s$.  
It serves as a terminal cost in the objective, discouraging trajectories that would lead to unsafe states beyond the planning horizon.
The value function is implemented as a small two-hidden-layer multi-layer perceptron (MLP) with 256 hidden units per layer and ReLU activations, trained in JAX \cite{jax2018github}.  
Layer normalization is applied after each hidden layer for training stability.

The initial training data is collected during optimization in simulation: at each planning cycle, the top $\rho = 0.3$ fraction of sampled trajectories (ranked by the total return $\mathcal{S}$) is saved for minimal offline training.
For every saved state $s_t$, we compute an $n$-step temporal difference (TD) target with $n = H$ (the planning horizon).  
Targets that would exceed the episode are truncated at the last observed step.  
This formulation ensures the learned value is aligned with the cost landscape that the planner actually optimizes.

The network is trained with the Adam optimizer at a learning rate of $3 \times 10^{-4}$, using gradient clipping at a global norm threshold of 100.0 and mean squared error loss.  
Target values are clipped to prevent outliers from dominating the loss.  
Training runs for $25\mathrm{k}$ epochs with a batch size of $8\mathrm{k}$.

\subsection{Hyperparameters}
\label{app:hyperparams}

\begin{table}[H]
\centering
\caption{Key hyperparameters for the safety filter pipeline.}
\label{tab:hyperparams}
\begin{tabular}{@{}lll@{}}
\toprule
\textbf{Category} & \textbf{Parameter} & \textbf{Value} \\
\midrule
\multirow{9}{*}{Replica Exchange} 
    & Samples ($K$) & 512 \\
    & Horizon ($H$) & 5 \\
    & Iterations ($N$) & 3 \\
    & Noise level & 0.6 \\
    & Discount ($\gamma$) & 0.99 \\
    & Replicas & 20 \\
    & $\lambda_{\min}$ / $\lambda_{\max}$ & 0.01 / 2.0 \\
    & Swap frequency & 1 iter \\
    & Momentum ($\beta$) & 0.1 \\
\midrule
\multirow{1}{*}{Geometric Projection} 
    & Clearance ($\varepsilon_{\text{safe}}$) & 0.08\,m \\
\midrule
\multirow{6}{*}{Value Network} 
    & Top fraction ($\rho$) & 0.3 \\
    & Hidden dim & 256 \\
    & Layers & 2 (MLP) \\
    & Learning rate & $3 \times 10^{-4}$ \\
    & Epochs & 25,000 \\
    & Batch size & 8,000 \\
\midrule
\multirow{4}{*}{Rollout Physics} 
    & Timestep & 0.03\,s \\
    & Solver iterations & 4 \\
    & Line search iterations & 4 \\
    & PD gains ($k_p, k_d$) & 30, 0.5 \\
\bottomrule
\end{tabular}
\end{table}

\section{Qualitative Evaluation}
\label{app:qualitative}

We demonstrate the filter’s behavior in three scenarios that highlight its core capabilities.

\begin{figure}[!htp]
  \centering
  \begin{subfigure}[t]{\textwidth}
    \centering
    \includegraphics[width=\linewidth]{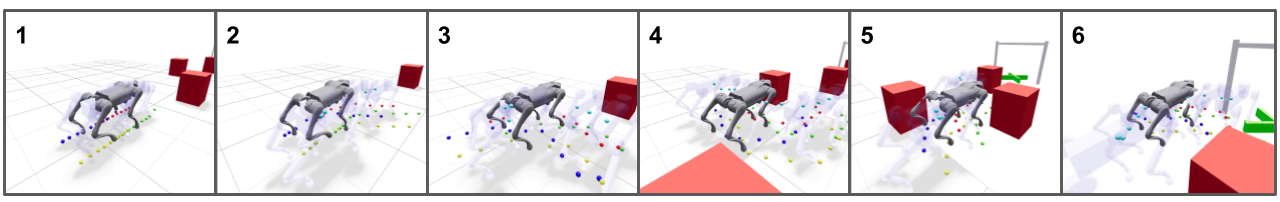}
    \caption{Large obstacle: the filter replans contacts (colored dots) to route around the obstacle. The transparent robot shows the planner's predicted trajectory.}
    \label{fig:sf_go2_nav}
  \end{subfigure}
  \vspace{4pt}
  \begin{subfigure}[t]{\textwidth}
    \centering
    \includegraphics[width=\linewidth]{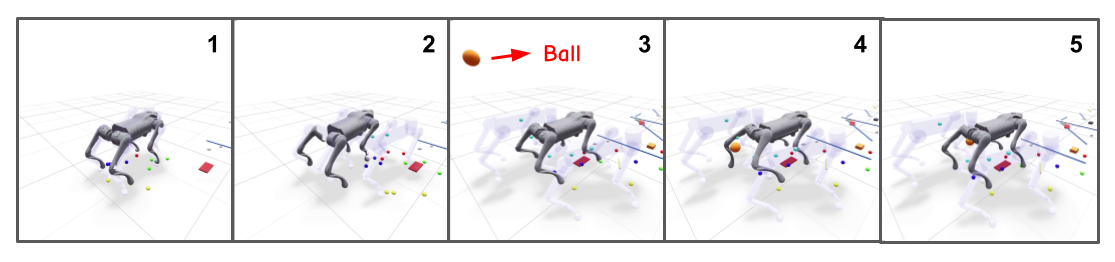}
    \caption{Dynamic obstacle: a ball approaches (frame 2) and the filter reactively shifts planned contacts to steer clear.}
    \label{fig:sf_go2_dynamic}
  \end{subfigure}
  \vspace{4pt}
  \begin{subfigure}[t]{\textwidth}
    \centering
    \includegraphics[width=\linewidth]{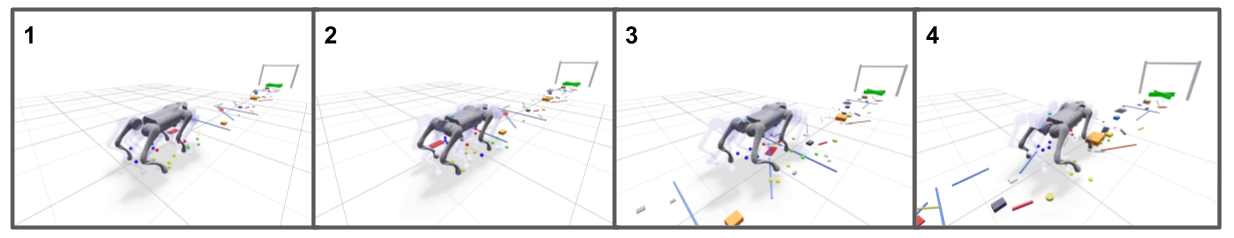}
    \caption{Cluttered scene (${\sim}50$ obstacles): the filter aggressively redirects contacts, trading path directness for collision avoidance.}
    \label{fig:sf_go2_cluttered}
  \end{subfigure}
  \caption{Qualitative results across three scenarios. Colored dots indicate planned foot-contact locations; the transparent robot shows the planner's predicted rollout. Keyframes progress left to right.}
  \label{fig:sf_go2_all}
\end{figure}

\paragraph{Scenario 1: Large obstacle navigation}
The robot must navigate past a large obstacle occluding a significant portion of the workspace.
As shown in Fig.~\ref{fig:sf_go2_nav}, the SDF cost guides the sampler toward contact sequences that route the robot around the obstacle, producing a smooth detour while maintaining a stable gait.

\paragraph{Scenario 2: Dynamic obstacle avoidance}
A moving obstacle crosses the robot's path.
Fig.~\ref{fig:sf_go2_dynamic} shows the filter modifying planned contacts on-the-fly to steer clear, demonstrating reactivity to time-varying constraints within the receding horizon.

\paragraph{Scenario 3: Cluttered environment}
Fig.~\ref{fig:sf_go2_cluttered} shows the filter finding safe contacts to carefully place the robot’s feet while avoiding multiple objects on the ground.

\end{document}